\documentclass[runningheads]{llncs}

\usepackage{eccv}

\usepackage{eccvabbrv}

\usepackage{graphicx}
\usepackage{booktabs}

\usepackage[accsupp]{axessibility}  %

\usepackage{hyperref}

\usepackage{orcidlink}

\usepackage[table]{xcolor}
\usepackage{collcell}
\usepackage{array}
\usepackage{multirow}
\usepackage{xfp}
\usepackage{tabularx}
\usepackage{arydshln}

\newcommand{\heatcell}[1]{%
  \ifdim #1pt > 75pt \cellcolor{green!50}#1%
  \else\ifdim #1pt > 50pt \cellcolor{green!30}#1%
  \else\ifdim #1pt > 25pt \cellcolor{green!15}#1%
  \else\ifdim #1pt > 0pt \cellcolor{green!5}#1%
  \else #1\fi\fi\fi\fi}

\begin{document}

\title{HistReNeRF: Historic Image Relocalisation within Contemporary Neural Radiance Field Reconstructions} 

\titlerunning{HistReNeRF}

\author{Benjamin T. Hughes \and
Stuart James\orcidlink{0000-0002-2649-2133}}

\authorrunning{B.~Hughes and S.~James}

\institute{Durham University, Durham, UK \\ \email{stuart.a.james@durham.ac.uk}}
\maketitle

\begin{abstract}
Relocalising archival photographs within a contemporary scene model is challenging because historic and modern views can differ in photographic appearance, visible objects, and spatial layout. Therefore, we present \textit{HistReNeRF}, a framework that estimates the 6-DoF pose of a historic photograph by matching adapted DINOv2 patch features to candidate rays sampled from a contemporary Neural Radiance Field (NeRF) reconstruction.  The continuous representation of a NeRF provides a queryable scene interface from which candidate rays can be sampled and matched, enabling domain adaptation between historic photography and contemporary images directly in the feature representation used for localisation. We evaluate embedding-space-based domain adaptation against pixel-space methods on a new cross-temporal dataset comprising 10,545 contemporary street-level images and 230 archival photographs from three European landmarks. Embedding-space adaptation reduces translation and rotation errors by an average of $11\%$ and $16\%$, respectively, across the three scenes. These results show that neural scene relocalisation provides a natural interface for feature-space adaptation, reducing cross-temporal appearance shift without modifying the query image. Code and dataset at \url{https://github.com/ARTUROLab/HistReNeRF}.

\end{abstract}

\section{Introduction}
Historic photographs provide an important record of past urban environments, architectural heritage, and visual culture. Relocalising such images within contemporary three-dimensional scene reconstructions would support the comparison of historical and present-day views, the documentation of urban change, and the spatial interpretation of archival collections. However, historic image relocalisation differs substantially from conventional visual localisation. Archival photographs may vary from contemporary imagery in photographic medium, film grain, contrast, resolution, and tonal range, while the depicted scene may have changed through demolition, renovation, vegetation growth, or the replacement of street furniture and surrounding structures. These combined appearance and structural changes make it difficult to establish reliable correspondences between a historic query image and a modern scene model.

Classical localisation pipelines typically estimate pose by matching local image descriptors, such as SIFT~\cite{Lowe2004}, to a Structure-from-Motion (SfM) reconstruction~\cite{schoenberger2016sfm}. Their success depends on repeatable local structure and sufficient photometric consistency between the query and reference images. Both assumptions are weakened in the historic setting where photographic changes affect the visibility and appearance of local features, while scene evolution can remove or alter the structures on which correspondences depend. Consequently, a contemporary point cloud may provide unreliable support for relocalising.

Neural scene representations offer an alternative reference model by coupling scene geometry and appearance within a continuous, queryable representation. A Neural Radiance Field (NeRF)~\cite{Nerf} models a scene as a volumetric function that can be rendered from arbitrary viewpoints, allowing localisation to be formulated against a learned scene model rather than a sparse set of 3D descriptors. Early relocalisation work, iNeRF~\cite{yenchen2021inerfinvertingneuralradiance}, used an inversion-based formulation, where it recovered camera pose by iteratively rendering a pretrained NeRF from candidate viewpoints and minimising its photometric difference from the query image. While this demonstrates the localisation potential of neural scenes, repeated rendering makes inference computationally expensive and dependent on a suitable pose initialisation. IFFNeRF~\cite{bortolon2024iffnerfinitialisationfreefast} instead formulates localisation through ray--query correspondence. Candidate rays sampled from the neural scene are matched to DINOv2~\cite{oquab2023dinov2} image features through cross-attention, and the pose is recovered by weighted least-squares ray minimum-distance estimation. This formulation is particularly relevant for historic imagery because it replaces direct low-level descriptor matching with a learned representation. However, historic queries remain separated from contemporary scene imagery by a substantial cross-temporal appearance gap.

To address the cross-temporal appearance gap, we present \textit{HistReNeRF}, a framework for relocalising historic photographs within contemporary TensoRF reconstructions. HistReNeRF matches DINOv2 patch features extracted from a historic query image to candidate rays sampled from the neural scene. We introduce an embedding-space CycleGAN that maps historic patch features towards the contemporary feature distribution before ray matching. 
We evaluate on a new cross-temporal dataset across three scenes: Arc de Triomphe, Brandenburg Gate, and Piazza del Duomo. 

The contributions of this paper are as follows: \textit{i)} We present \textit{HistReNeRF}, a cross-temporal neural scene relocalisation framework that applies unpaired domain adaptation directly to the query representation used for ray matching, and compares feature-space transfer against pixel-space translation. \textit{ii)} We introduce a new dataset for historic image relocalisation with validated reference poses for historic queries across three European landmark scenes. \textit{iii)} We provide a systematic evaluation of embedding-space and pixel-space adaptation for cross-temporal relocalisation, showing that feature-space transfer yields consistent gains.

\section{Related Work}
We review scene reconstruction techniques in Secs.~\ref{sec:rel_sec1} and relocalisation in \ref{sec:rel_relocalisation},  followed by domain adaptation methods addressing the appearance gap between historic and contemporary imagery in
Sec.~\ref{sec:rel_domain_adaptation}. Finally, we review cross-domain and localisation approaches in Sec.~\ref{sec:rel_datasets}.

\subsection{Scene Reconstruction}
\label{sec:rel_sec1}
Classical Structure-from-Motion methods reconstruct camera poses and sparse three-dimensional point clouds from overlapping image collections, establishing the geometric basis for large-scale landmark reconstruction from crowd-sourced imagery~\cite{10.1145/1141911.1141964,RomeInADay}. These reconstructions support conventional localisation through image-to-point correspondences, but their sparse structure provides limited support when the query differs substantially in appearance from the images used to build the model.
Similarly, 3D Gaussian Splatting represents the scene through an explicit set of optimised Gaussian primitives and achieves real-time rendering through differentiable rasterisation~\cite{kerbl20233dgaussiansplattingrealtime}. 

Alternatively, neural scene representations extend this geometric reference with a dense model of scene appearance. Neural Radiance Fields (NeRFs) represent a scene as a volumetric function that maps position and viewing direction to density and colour, enabling novel views to be synthesised from arbitrary camera poses~\cite{Nerf}. Subsequent work has made these representations more practical for real scenes primarily focusing on improving the speed of learning the function. TensoRF~\cite{chen2022tensorf} factorises the radiance field into compact tensor components, substantially reducing the computational cost of reconstruction and rendering. In contrast, Instant-NGP~\cite{muller2022instant} accelerated neural-field optimisation through multiresolution hash encodings. Extensive surveys have been conducted on NeRFs~\cite{xie2022neural}; however, beyond generalising to the wild with unconstrained contemporary camera imagery~\cite{martin2021nerf}, archival photography remains under-explored.

\subsection{Visual Relocalisation in Neural Scenes}
\label{sec:rel_relocalisation}

Classical visual relocalisation estimates the 6-DoF camera pose of a query image by matching local features against a three-dimensional reconstruction. Hand-crafted descriptors such as SIFT~\cite{Lowe2004} and learned alternatives such as SuperPoint~\cite{detone2018superpoint} can provide correspondences when the query and reference imagery share sufficient local structure. 
Beyond correspondence-based approaches, diffusion models have been explored for probabilistic camera pose estimation, either by modelling distributions over camera parameters~\cite{wang2023posediffusion} or over distributed ray-based camera representations~\cite{zhang2024raydiffusion}.
However, correspondence-based relocalisation remains sensitive to appearance changes that alter the visibility, texture, or contrast of local regions, making it difficult to apply directly to archival photographs.

Neural scene representations have introduced alternatives to descriptor-based localisation. iNeRF~\cite{yenchen2021inerfinvertingneuralradiance} recovers pose by iteratively rendering a pretrained NeRF from candidate viewpoints and minimising photometric error against the query image. Alternatively, VRS-NeRF~\cite{xue2024vrsnerfvisualrelocalizationsparse} reduces the cost of this analysis-by-synthesis formulation for sparse neural scenes, while retaining its dependence on optimisation against the rendered scene.

A complementary direction uses dense learned features from neural scene representations to establish correspondences more directly. CrossFire~\cite{moreau2023crossfirecamerarelocalizationselfsupervised} derives self-supervised features from an implicit scene representation for camera relocalisation, while pNeRF-Loc~\cite{zhao2023pnerflocvisuallocalizationpointbased} and NeRFect Match~\cite{nerfectmatch2024} investigate the use of NeRF-based representations for visual localisation and correspondence estimation. 
IFFNeRF~\cite{bortolon2024iffnerfinitialisationfreefast} develops this direction into a feed-forward formulation that matches query-image features to candidate rays sampled from a neural scene and recovers pose without iterative inversion. 
Overall, these methods show that neural scenes can support localisation through learned correspondence representations, but they generally assume that query images remain visually compatible with the imagery used to reconstruct the scene.

\subsection{Unpaired Domain Adaptation}
\label{sec:rel_domain_adaptation}
Domain adaptation addresses the problem of applying a model across visual domains with different appearance distributions. In the unpaired setting, i.e. without one-to-one correspondences, CycleGAN learns mappings between source and target domains through adversarial training and cycle consistency, without requiring corresponding image pairs~\cite{zhu2017unpaired}. This is particularly relevant to historic imagery, where equivalent historic and contemporary photographs are rarely available from the same 6DoF viewpoint.

Relaxing the cycle consistency,  methods such as UNIT~\cite{liu2017unsupervisedImageToImage} model the two domains through a shared latent representation, while MUNIT~\cite{huang2018multimodalunsupervisedimage} separates content from style to support multiple plausible translations from a common source image. CUT~\cite{park2020cut} instead uses patch-wise contrastive learning to preserve local content while learning a one-directional translation between unpaired domains. These methods provide different mechanisms for altering image appearance while preserving the scene structure required by a downstream task.

Pixel-space adaptation has been used to improve performance under domain shift in perception and localisation. Style transfer has reduced the synthetic-to-real gap for monocular depth estimation, while night-to-day translation has been explored for retrieval-based visual localisation~\cite{atapourBreckon,anoosheh2019nighttodayimagetranslationretrievalbased}. In each case, adaptation is applied to pixels before downstream features are extracted. This creates a potential limitation for historic image relocalisation, where a translation model must alter photographic appearance without changing the architectural and spatial evidence required for pose estimation.

\subsection{Domain Adaptation and Historic Visual Data}
\label{sec:rel_datasets}

Visual localisation benchmarks have increasingly examined changing environmental conditions. Cambridge Landmarks~\cite{kendall2015posenet} provides outdoor landmark scenes for camera-pose estimation, while Aachen Day-Night ~\cite{sattler2018benchmarking} evaluates localisation across pronounced illumination changes and RobotCar Seasons extends this setting across weather and seasonal variation~\cite{maddern20171}. These benchmarks are important for evaluating appearance robustness, but they retain a relatively stable photographic process and scene structure compared with the century-scale change encountered in archival imagery.

Historic visual datasets have addressed related but distinct problems. Maiwald~\cite{maiwald2019benchmark} constructs a benchmark for evaluating feature matching on historical architectural photographs, directly exposing the difficulty of establishing correspondences across temporal appearance change. Komorowicz et al.~\cite{komorowicz2024coloringpast} use archival photographs for neural reconstruction of historical monuments, addressing degraded imagery and incomplete visual evidence from the historical domain. These works demonstrate the importance of historic imagery for geometric and visual analysis, but they do not consider the task of localising an archival photo against a separately reconstructed contemporary neural scene, which we address.

\section{HistReNeRF}
\label{sec:methodology}
\begin{figure*}
    
    \centering
    \includegraphics[width=1\linewidth]{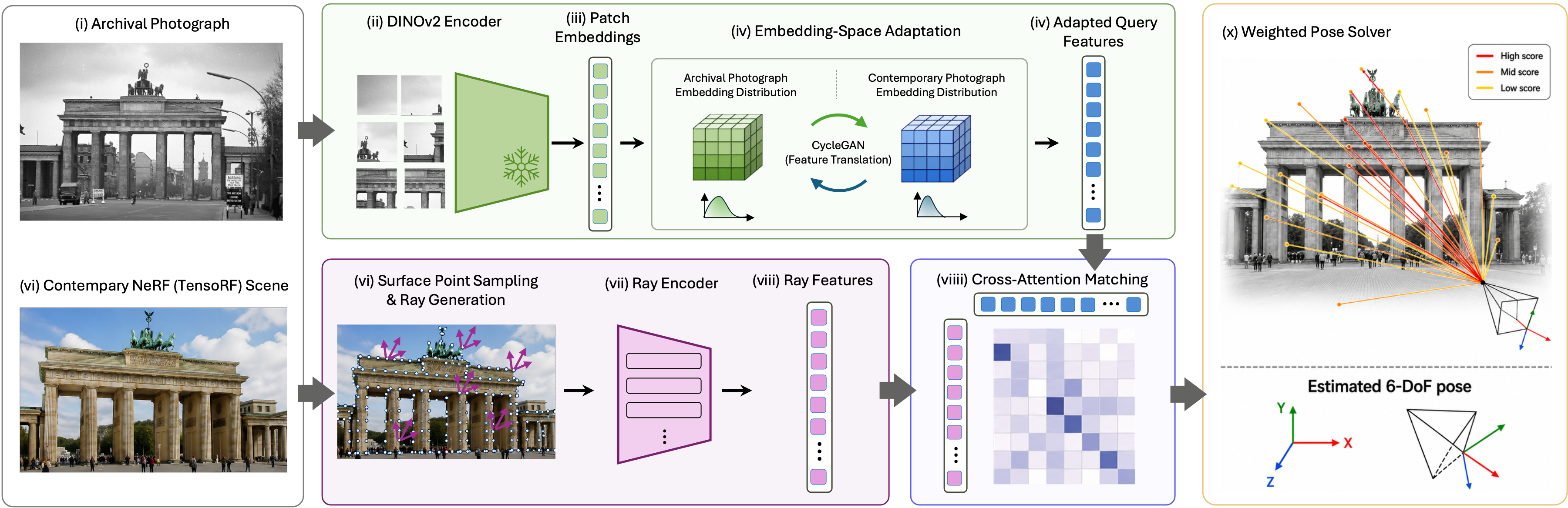}
    \caption{\textbf{Overview of \textit{HistReNeRF}.} Given a historic query image (i), it is encoded by frozen DINOv2 (ii) into patch embeddings (iii), which are mapped towards the contemporary embedding distribution by embedding-space adaptation (iv) to produce adapted query features (v). In parallel, a contemporary TensoRF reconstruction (vi) is used to sample candidate surface points and generate normal-oriented candidate rays, and encode them as ray features (viii). Cross-attention matching (viiii) scores the compatibility between adapted query features and candidate rays. The weighted pose solver (x) aggregates the highest-scoring rays to estimate the 6-DoF camera pose of the historic photograph within the contemporary neural scene.}

\label{fig:pipeline}
\end{figure*}

Given a historic query photograph $I_q$ and a contemporary TensoRF neural radiance field reconstruction $f_\theta$ of the same scene, HistReNeRF estimates the 6-DoF camera pose $\hat{\mathbf{P}} = (\hat{\mathbf{R}}, \hat{\mathbf{t}})$ in the coordinate frame of the reconstruction. The method uses an IFFNeRF-style ray--query localiser trained on contemporary imagery. For the ray--query localiser, candidate rays are sampled from the TensoRF density field and matched to DINOv2 patch features extracted from the historic query through cross-attention. The highest-scoring rays estimate the camera centre $\hat{\mathbf{t}}$ through a weighted least-squares minimum-distance estimate, while their directions and the scene up-vector determine the camera orientation $\hat{\mathbf{R}}$. To reduce the historic-to-modern appearance gap, HistReNeRF applies unpaired domain adaptation directly to the DINOv2 patch representation used for ray--query matching. A CycleGAN maps historic patch embeddings towards the distribution of embeddings extracted from selected contemporary images after feature extraction and before cross-attention matching. The DINOv2 backbone, TensoRF reconstruction, ray--query localiser, and historic query image remain unchanged throughout adaptation. Figure~\ref{fig:pipeline} provides an overview of the pipeline.

\subsection{Neural Scene Relocalisation}
\label{sec:method_relocalisation}

We adopt the candidate-ray relocalisation formulation of IFFNeRF~\cite{bortolon2024iffnerfinitialisationfreefast}. Given a query image and a contemporary TensoRF reconstruction, samples of likely surface locations are extracted from the neural scene; from these point locations, multiple rays are cast. The query image is matched against this ray set, allowing the most relevant rays to be identified and combined into a pose estimate.

\paragraph{Surface-point sampling.}
We first extract $G$ candidate ray origins from the density field of the contemporary TensoRF reconstruction. Let $\hat{\sigma}(\mathbf{x})$ denote the density predicted at position $\mathbf{x}$. The procedure is initialised by uniformly sampling $G$ points within the scene bounding box. These points are then refined through an iterative Metropolis--Hastings procedure. At iteration $i$, a new random position is proposed for each point. The proposal for point $g$ is accepted only when its predicted density satisfies $\hat{\sigma}(\mathbf{u}_{g}^{i+1}) \geq \tau,$ where the density threshold $\tau$ is determined from the empirical cumulative density distribution of the selected points in the preceding iteration. Specifically, the procedure retains proposals associated with the upper portion of this distribution, using the $60$th-percentile criterion. The points retained after the final iteration form the set $\mathcal{U} = \{\mathbf{u}_g\}_{g=1}^{G}$, where $g$ indexes the $G$ sampled points; these provide the ray origins for the following ray-generation stage.

\paragraph{Ray generation.}
Each sampled point provides the origin for a set of rays. The neural scene is queried for the surface normal by the density gradient at $\mathbf{u}_g$, and ray directions are oriented by this normal to avoid casting rays towards the unobserved interior of the reconstructed surface. The deterministic isocell distribution~\cite{masset2011,beckers2016} is applied, which partitions the unit sphere into equal-area cells and uses their centres to obtain approximately uniform directions, where $V = 27$ rays are generated for each surface point, as in IFFNeRF. Each ray is represented by its origin $\mathbf{o}_j$, direction $\mathbf{d}_j$, and rendered colour $\mathbf{c}_j$. The colour is obtained by volumetric rendering along the ray within the TensoRF reconstruction. The resulting set contains $N = GV$ candidate rays
\begin{equation}
\mathcal{R} =
\left\{
\left(
\mathbf{o}_j,
\mathbf{d}_j,
\mathbf{c}_j
\right)
\right\}_{j=1}^{N}.
\label{eq:ray_set}
\end{equation}
This ray bundle spans possible camera locations and viewing directions within the reconstructed scene.

\paragraph{Ray--query matching.}
The historic query image $I_q$ is encoded using a frozen DINOv2 ViT-S/14 backbone~\cite{oquab2023dinov2}. For a $224 \times 224$ input image, the encoder produces $\mathbf{F}_q = \phi(I_q) \in \mathbb{R}^{K \times 384},$ where $K = 256$. The geometric and appearance parameters of each candidate ray are encoded separately. A ray encoder applies positional encoding to the ray origin, direction, and rendered colour before mapping them to ray features, $\mathbf{F}_r = \psi_r(\mathcal{R}) \in \mathbb{R}^{N \times d}.$
Positional encoding expands small differences in ray origin and direction, allowing the encoder to distinguish rays with similar rendered appearance but different geometric configurations.

The query-image features act as attention queries and the ray features act as attention keys. Their compatibility is represented by an attention matrix
\begin{equation}
\mathbf{A}
\left(
\mathbf{F}_q,
\mathbf{F}_r
\right)
\in
\mathbb{R}^{K \times N},
\label{eq:attention}
\end{equation}
where $\mathbf{A}$ denotes the learned attention module. Each entry $\mathbf{A}_{k,j}$ measures the compatibility between query-image patch $k$ and candidate ray $j$. Summing attention over the image patches gives a score for each candidate ray
\begin{equation}
\hat{s}_j =
\sum_{k=1}^{K}
A_{k,j}.
\label{eq:ray_score}
\end{equation}

\paragraph{Geometric supervision.}
The ray scorer is trained on contemporary images with known camera centres. For a training image with camera centre $\mathbf{t}^{\ast}$, the supervision signal measures how closely each candidate ray passes to that camera centre. The closest point on ray $j$ is $\mathbf{p}_j = \mathbf{o}_j + \max \left( (\mathbf{t}^{\ast} - \mathbf{o}_j)^{\top} \mathbf{d}_j, 0 \right) \mathbf{d}_j,$ where the maximum restricts the projection to the forward ray half-line. The distance between the ray and the camera centre is then $\delta_j = \left| \mathbf{p}_j - \mathbf{t}^{\ast} \right|_2.$ This distance is converted into a target score by
\begin{equation}
\tilde{s}_j =
1 -
\tanh
\left(
\frac{
\delta_j
}{
\lambda
}
\right),
\label{eq:unnormalised_target_score}
\end{equation}
where $\lambda$ controls the range of target scores.
Following IFFNeRF, the scores are normalised to sum to K, matching the scale of the aggregated attention scores:
\begin{equation}
s_j =
\frac{
K \tilde{s}_j
}{
\sum_{\ell=1}^{N}
\tilde{s}_\ell
}.
\label{eq:target_score}
\end{equation}
The ray scorer is trained by minimising the $\ell_2$ discrepancy between predicted and target scores.

\paragraph{Pose recovery.}
At inference, the highest-scoring rays define a set $\mathcal{J}$. The estimated camera centre is the weighted least-squares minimum-distance estimate of these rays
\begin{equation}
\hat{\mathbf{t}} =
\left(
\sum_{j \in \mathcal{J}}
\hat{s}_j
\left(
\mathbf{I}_3 -
\mathbf{d}_j
\mathbf{d}_j^{\top}
\right)
\right)^{-1}
\sum_{j \in \mathcal{J}}
\hat{s}_j
\left(
\mathbf{I}_3 -
\mathbf{d}_j
\mathbf{d}_j^{\top}
\right)
\mathbf{o}_j.
\label{eq:camera_centre}
\end{equation}
The selected ray bundle provides the corresponding camera-pose estimate within the TensoRF coordinate frame. HistReNeRF retains this localiser unchanged and adapts only the historic DINOv2 representation before ray--query matching.

\subsection{Contemporary Domain Relevance Filtering}
\label{sec:method_domain_selection}

Contemporary imagery used to reconstruct the neural scene often consists of dash-cam or street-level views, including images dominated by roads, vehicles, or foreground street furniture despite being geographically aligned with the landmark and reconstruction. While these images provide useful geometric coverage for TensoRF reconstruction, many are unsuitable targets for cross-temporal adaptation. Such views can differ substantially from the architectural content represented in the historic collection. Training a domain-transfer model on the full contemporary image set may therefore encourage appearance changes that are unrelated to the historic-to-modern shift relevant for relocalisation.

We select a contemporary target subset using global CLIP image embeddings~\cite{radford2021clip}. CLIP filtering is used only for within-scene view selection for domain adaptation, rather than for landmark identification or camera localisation. Let $\mathbf{e}^{h}_i$ denote the embedding of historic training image $I^{h}_i$ and let $\mathbf{e}^{m}$ denote the embedding of a candidate contemporary image $I^{m}$. A contemporary image is retained when it lies sufficiently close to at least one historic image in CLIP embedding space. Specifically, its nearest historic embedding must be no more than $1.5$ times the maximum pairwise cosine distance observed between images in the historic collection. This defines a scene-specific threshold that excludes visually dissimilar contemporary views while allowing for the expected historic-to-modern appearance shift.

\begin{equation}
\min_i d*{\cos}(\mathbf{e}^{m}, \mathbf{e}^{h}_i)
\leq
\alpha
\max{i,j}
d*{\cos}(\mathbf{e}^{h}_i, \mathbf{e}^{h}_j),
\label{eq:contemporary_selection}
\end{equation}
where $d*{\cos}$ is cosine distance and $\alpha$ controls the permitted historic-to-modern variation. This selection retains contemporary views that are visually compatible with the architectural content of the historic collection while preserving sufficient variation to represent the contemporary domain. The selected images define the target distribution used to train the embedding-space adaptation model.

\subsection{Embedding-Space Adaptation}
\label{sec:method_embedding_adaptation}

The contemporary subset selected in Sec.~\ref{sec:method_domain_selection} defines the target visual distribution for adaptation. Rather than modifying the historical photograph itself, HistReNeRF aligns the DINOv2 representation used by the ray--query matching module. Given an image $I$, the frozen DINOv2 encoder encodes the image expressed as $\mathbf{E} = \phi(I) \in \mathbb{R}^{256 \times 384}$, which can be reshaped to $16 \times 16 \times 384$ to preserve the original patch layout of the image. This representation retains the spatial organisation of the historic query while expressing its content in the semantic feature space used by the localisation model.

Let $\mathbf{E}_h \sim \mathcal{D}_h$ and $\mathbf{E}_m \sim \mathcal{D}_m$ denote embedding grids extracted from historic and selected contemporary images. We train an unpaired CycleGAN~\cite{zhu2017unpaired} to learn mappings between these distributions. The model contains two U-Net-style convolutional generators, $G_{h \rightarrow m}$ and $G_{m \rightarrow h}$, together with two PatchGAN discriminators, $D_m$ and $D_h$. The generators operate over the spatial embedding grid, allowing the translation to account for local feature structure while maintaining the patch arrangement inherited from the query image.

The model is trained with least-squares adversarial losses~\cite{mao2017lsgan}, cycle consistency, and identity preservation. The complete objective is
\begin{equation}
\mathcal{L}_{\mathrm{emb}} =
\mathcal{L}_{\mathrm{adv}} +
\lambda_{\mathrm{cyc}} \mathcal{L}_{\mathrm{cyc}} +
\lambda_{\mathrm{id}} \mathcal{L}_{\mathrm{id}}.
\label{eq:embedding_loss}
\end{equation}
Here, $\mathcal{L}_{\mathrm{adv}}$ follows the standard least-squares GAN objective and aligns translated historic embeddings with the contemporary embedding distribution. Cycle consistency encourages a translated representation to remain recoverable after mapping back to its original domain
\begin{multline}
\mathcal{L}_{\mathrm{cyc}} =
\mathbb{E}_{\mathbf{E}_h \sim \mathcal{D}_h}
[
\|G_{m \rightarrow h}(G_{h \rightarrow m}(\mathbf{E}_h)) - \mathbf{E}_h\|_1
]
\\
+
\mathbb{E}_{\mathbf{E}_m \sim \mathcal{D}_m}
[
\|G_{h \rightarrow m}(G_{m \rightarrow h}(\mathbf{E}_m)) - \mathbf{E}_m\|_1
].
\label{eq:embedding_cycle}
\end{multline}

The identity term penalises unnecessary changes when an embedding is already sampled from the target domain
\begin{multline}
\mathcal{L}_{\mathrm{id}} =
\mathbb{E}_{\mathbf{E}_m \sim \mathcal{D}_m}
[
\|G_{h \rightarrow m}(\mathbf{E}_m) - \mathbf{E}_m\|_1
]
\\
+
\mathbb{E}_{\mathbf{E}_h \sim \mathcal{D}_h}
[
\|G_{m \rightarrow h}(\mathbf{E}_h) - \mathbf{E}_h\|_1
].
\label{eq:embedding_identity}
\end{multline}

We use $\lambda_{\mathrm{cyc}} = 10$ and $\lambda_{\mathrm{id}} = 5$. The DINOv2 encoder and neural-scene relocalisation model remain frozen throughout training, ensuring that the learned mapping aligns the historic representation with the existing contemporary localisation space rather than changing the representation used by the localiser.

At inference time, a historic query image is encoded by DINOv2, translated by $G_{h \rightarrow m}$, and restored to the patch-token representation.
The adapted features $\widetilde{\mathbf{F}}_q$ replace $\mathbf{F}_q$ in the cross-attention module described in Sec.~\ref{sec:method_relocalisation}. The historic image, the contemporary TensoRF reconstruction, and the ray-based pose solver are not modified.

\section{Evaluation}
\label{sec:eval}

We evaluate HistReNeRF on cross-temporal relocalisation within contemporary neural scene reconstructions. Our experiments assess whether embedding-space adaptation improves pose estimation for historic query images, how it compares with pixel-space translation methods, and how performance varies across scenes with different viewpoint and appearance gaps. We first introduce the \textit{HistScene} dataset in Sec.~\ref{sec:dataset}, before reporting quantitative and qualitative comparisons between no adaptation, pixel-space adaptation, and the proposed embedding-space approach in Sec.~\ref{sec:eval_relocalisation}.

\subsection{HistScene Dataset}
\label{sec:dataset}

The HistScene dataset is a cross-temporal relocalisation dataset comprising 10,545 contemporary street-level images and 230 archival photographs across three locations: Arc de Triomphe in Paris, Brandenburg Gate in Berlin, and Piazza del Duomo in Milan. The three scenes provide complementary challenges. Arc de Triomphe exhibits substantial viewpoint mismatch between historic and contemporary imagery and strong architectural symmetry. Brandenburg Gate contains predominantly ground-level views with more comparable historic and contemporary camera distributions. Piazza del Duomo provides a wider architectural context that includes the cathedral facade and surrounding piazza.

For each scene, contemporary Mapillary imagery is used to recover a COLMAP reconstruction and train a TensoRF scene representation. Historic photographs were collected from Europeana and Wikimedia Commons and span a substantial range of photographic processes, image conditions, viewpoints, and scene content. Selected images were based on their Creative Commons licence. The historical collection is divided based on whether a reliable reference pose can be recovered in the contemporary COLMAP reconstruction. Successfully registered images, as explained below, form the evaluation set, while images without validated registration serve as unpaired historical training samples for domain adaptation. Dataset statistics are reported in Table~\ref{tab:dataset_stats}.

\begin{table}[t]
\scriptsize
\centering
\caption{HistScene statistics. Pose-validated images form the historic evaluation set. Remaining historic photographs are used as unpaired samples for domain adaptation training.}
\label{tab:dataset_stats}
\begin{tabularx}{\textwidth}{Xcccc}
\hline
Scene & Contemporary & Historic & Pose-validated (Test) & Training \\
\hline
Arc de Triomphe & 2,371 & 92 & 25 & 67 \\
Brandenburg Gate & 5,000 & 96 & 18 & 78 \\
Piazza del Duomo & 3,174 & 42 & 15 & 27 \\
\hline
Total & 10,545 & 230 & 58 & 172 \\
\hline
\end{tabularx}
\end{table}

\begin{table}[t]
\centering
\caption{
Reference-pose validation metrics averaged per scene. Gradient NCC and Edge F1 assess structural alignment between TensoRF renders and query images; higher is better. Grayscale SSIM is a photometric sanity check and is expected to be lower for historic images.
}
\label{tab:gt_validation}
\scriptsize
\begin{tabularx}{\textwidth}{Xcccccc}
\hline
 & \multicolumn{2}{c}{Gradient NCC} & \multicolumn{2}{c}{Edge F1} & \multicolumn{2}{c}{SSIM (gray)} \\
\cline{2-7}
Scene & Modern & Historic & Modern & Historic & Modern & Historic \\
\hline
Arc de Triomphe  & 0.124 & 0.047 & 0.240 & 0.637 & 0.499 & 0.268 \\
Brandenburg Gate & 0.044 & -0.001 & 0.116 & 0.217 & 0.375 & 0.351 \\
Piazza del Duomo & 0.089 & -0.004 & 0.092 & 0.099 & 0.413 & 0.418 \\
\hline
\end{tabularx}
\end{table}
\textbf{Ground Truth Pose Verification.} 
We recover reference poses for historic images by registering them against the existing COLMAP sparse reconstruction. SIFT correspondences between the historic query and reconstructed three-dimensional points are used within PnP-RANSAC to estimate camera pose and intrinsics. A query is retained for quantitative pose evaluation only when registration succeeds and a manual inspection confirms that reprojected scene points align with visible architectural structure. To achieve this, sparse points are reprojected onto a neural-scene rendering from the COLMAP camera pose. This verification is performed independently of HistReNeRF predictions.

Table~\ref{tab:gt_validation} reports gradient-normalised cross-correlation, edge F1, and grayscale SSIM between each historic query and a TensoRF rendering from its recovered pose. These quantities are used as supporting structural diagnostics rather than pose-selection criteria, since cross-temporal differences in photographic appearance make photometric agreement inherently unreliable. Together, geometric reprojection and visual inspection provide the basis for the validated reference poses used in the following experiments.

\begin{figure}[t]
    \centering
    \includegraphics[width=0.49\linewidth]{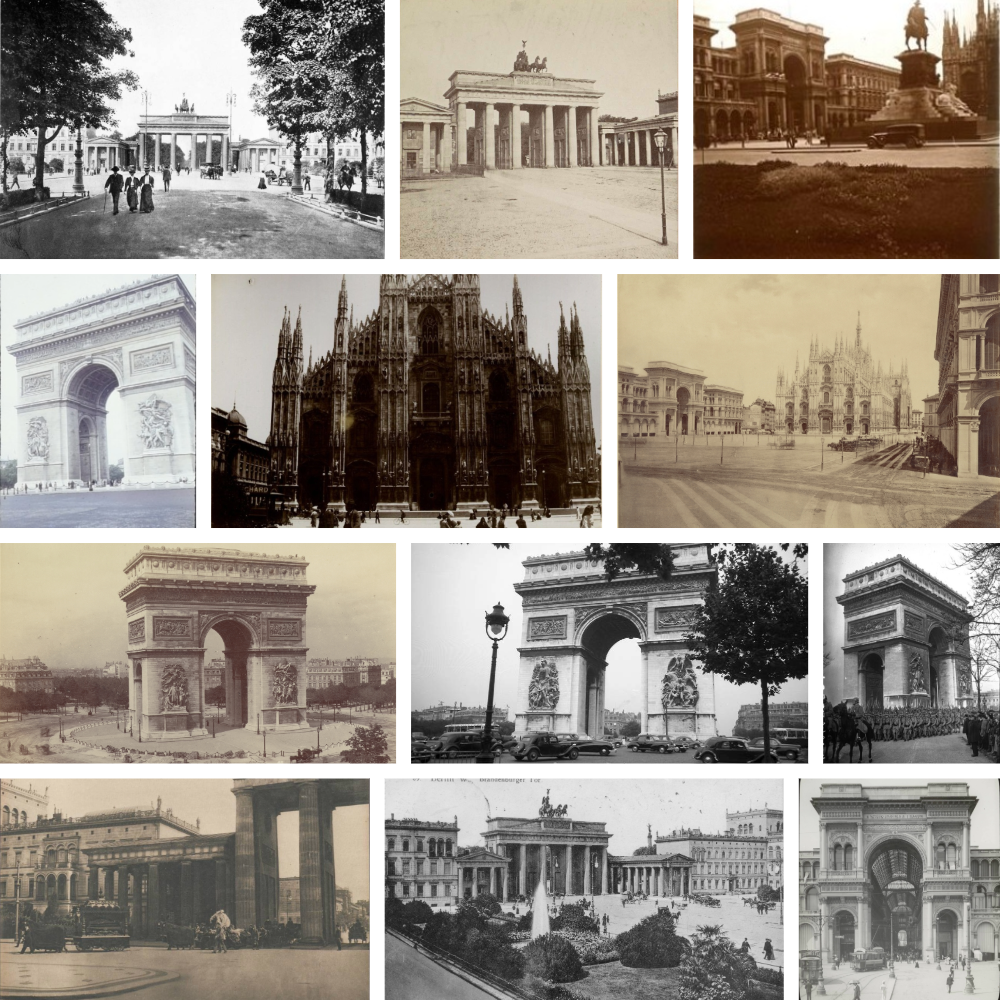}
    \includegraphics[width=0.49\linewidth]{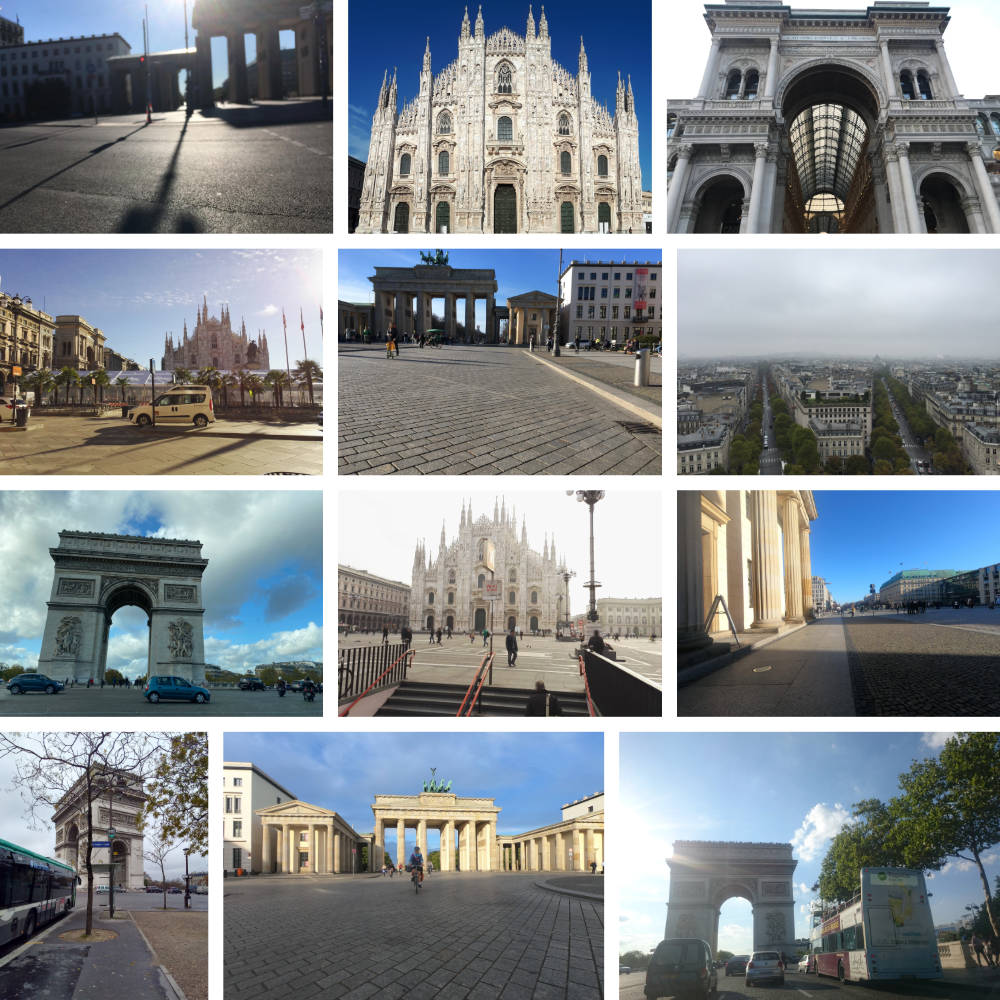}
    \caption{Example images from the proposed HistScene dataset facing the monument with left archival photography and right Mapillary street view.}
    \label{fig:placeholder}
\end{figure}

\subsection{Cross-Temporal Relocalisation Evaluation}
\label{sec:eval_relocalisation}
We evaluate cross-temporal relocalisation on the 58 pose-validated historic queries in HistScene. We report mean translation error, MTE, in metres and mean angular error, MAE, in degrees, where lower values indicate more accurate pose estimation. All configurations use the same contemporary TensoRF reconstruction, candidate-ray cache, frozen DINOv2 backbone, ray--query matching module, and pose solver. The comparison therefore isolates the representation at which cross-temporal adaptation is applied. We compare the following configurations as well as the proposed HistReNeRF:
\begin{itemize}
\item \textbf{No adaptation:} DINOv2 patch features are extracted directly from the historic query and matched to candidate rays from the contemporary neural scene.

\item \textbf{Img.\ CycleGAN:} Translates the historic query image towards the contemporary domain using adversarial training and cycle consistency~\cite{zhu2017unpaired}. DINOv2 features are then extracted from the translated image.

\item \textbf{Img.\ CUT:} Translates the historic image through a one-directional contrastive objective that encourages preservation of local image patches~\cite{park2020cut}.

\item \textbf{Img.\ UNIT:} Translates the historic image through a shared latent representation learned across historic and contemporary image domains~\cite{liu2017unsupervisedImageToImage}.

\item \textbf{Img.\ MUNIT:} Decomposes image content and style to generate plausible contemporary-domain translations from the historic query~\cite{huang2018multimodalunsupervisedimage}.

\end{itemize}

Each adaptation model is trained separately for each scene using the same historic training images and selected contemporary target images. Apart from the adaptation stage, all configurations share the same neural scene, feature encoder, ray--query matching module, and pose solver. We additionally report iNeRF refinement initialised from the ray-based pose estimate.

\begin{figure}[t!]
    \centering
    \includegraphics[width=0.99\linewidth]{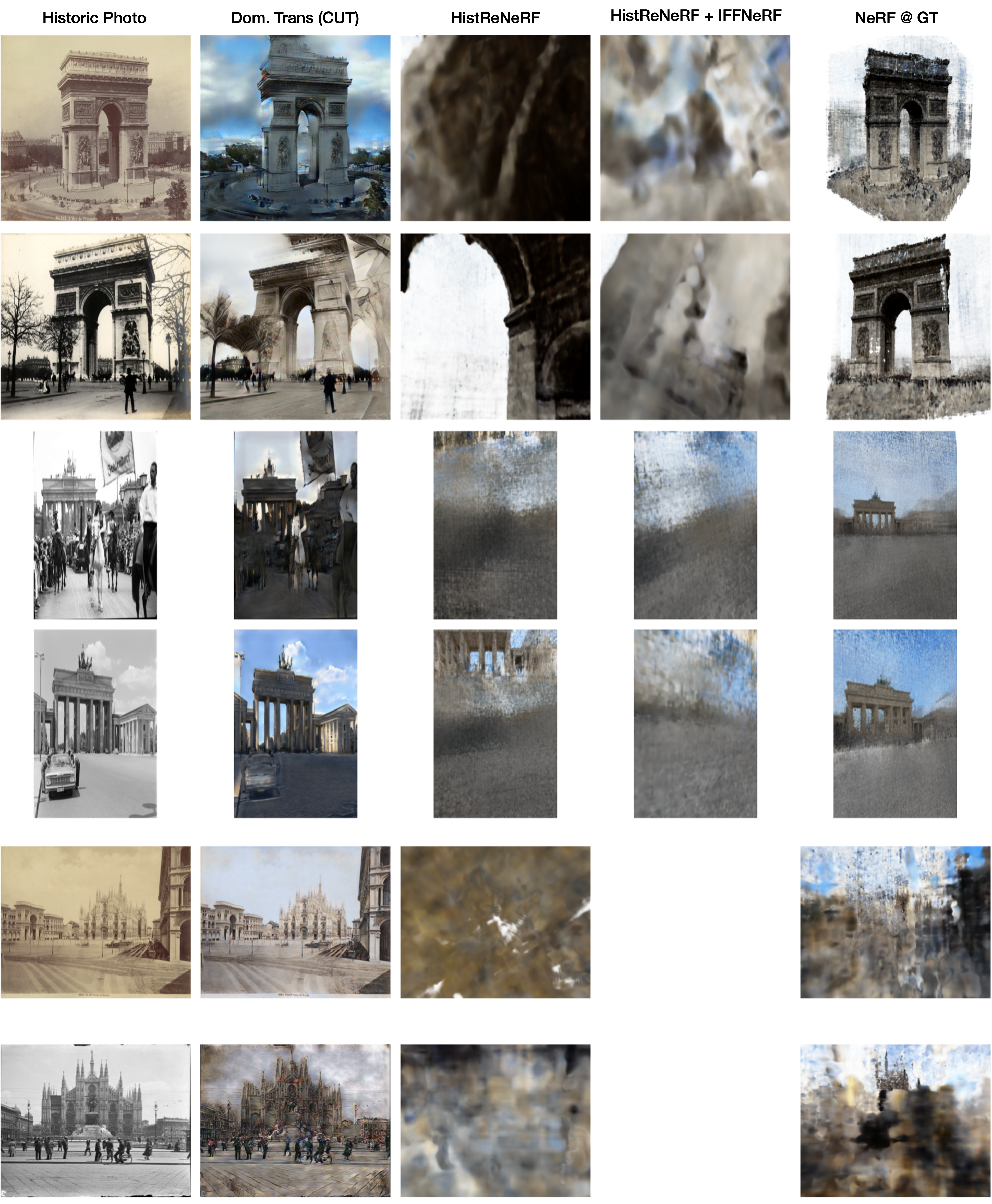}
    \caption{Qualitative results across scenes. Each row shows a historic query image (col.\ 1), the domain-transferred output from the best pixel-space translation method (col.\ 2), the novel view rendered by TensoRF at the HistReNeRF estimated pose (col.\ 3), the render at the HistReNeRF + iNeRF refined pose (col.\ 4), and the ground-truth render at the registered pose (col.\ 5). Misalignment between cols.\ 3--4 and col.\ 5 reflects localisation error. The renders are visually imprecise across all methods, consistent with the high MTE and MAE values reported in Table~\ref{tab:results_all}.}
    \label{fig:qualitative_results}
    \vspace{20pt}
\end{figure}

\begin{table}[t!]
\scriptsize
\centering
\caption{HistScene relocalisation  MTE is reported in metres and MAE in degrees (lower is better). Best result per scene and metric is shown in bold.}
\label{tab:results_all}
\begin{tabularx}{\textwidth}{Xccccccc}
\hline
 & &  \multicolumn{2}{c}{Arc} & \multicolumn{2}{c}{Brandenburg} & \multicolumn{2}{c}{Duomo} \\
\cline{3-8}
Method & Setting &  MTE & MAE & MTE & MAE & MTE & MAE \\
\hline
IFFNeRF               & \textit{Historic --- no adaptation}      & 6.624 & 71.7° & 4.712 & 23.4° & 1.783 & 85.5° \\
IFFNeRF + iNeRF       &\textit{Historic --- no adaptation}      & 6.818 & 81.5° & 5.279 & 35.7° & 2.234 & 71.0° \\
\hline
HistReNeRF         &\textit{Historic --- domain transfer}      & \textbf{6.608} & \textbf{69.6°} & \textbf{3.912} & \textbf{23.4°} & \textbf{1.589} & \textbf{58.6°} \\
HistReNeRF + iNeRF  &\textit{Historic --- domain transfer}     & 6.608 & 76.4° & 4.708 & 29.2° & 1.902 & 68.9° \\ \hdashline 
Img. CycleGAN        &\textit{Historic --- domain transfer}       & 7.179 & 88.6°  & 4.809 & 25.6° & 1.799 & 72.3° \\
Img. CycleGAN + iNeRF &\textit{Historic --- domain transfer}      & 6.966 & 103.5° & 5.703 & 42.6° & 1.973 & 64.0° \\
Img. CUT   &\textit{Historic --- domain transfer}                 & 7.007 & 90.9°  & 4.543 & 28.2° & 1.812 & 80.7° \\
Img. CUT + iNeRF  &\textit{Historic --- domain transfer}          & 7.079 & 106.1° & 5.323 & 38.8° & 1.961 & 90.9° \\
Img. UNIT     &\textit{Historic --- domain transfer}              & 7.337 & 112.5° & 5.427 & 30.8° & 1.984 & 85.0° \\
Img. UNIT + iNeRF &\textit{Historic --- domain transfer}          & 7.483 & 111.0° & 6.108 & 39.6° & 2.062 & 80.7° \\
Img. MUNIT  &\textit{Historic --- domain transfer}                & 8.548 & 118.1° & 7.608 & 33.9° & 3.071 & 121.8° \\
Img. MUNIT + iNeRF   &\textit{Historic --- domain transfer}       & 8.686 & 124.7° & 7.441 & 38.3° & 3.238 & 119.1° \\
\hline
\end{tabularx}
\end{table}

\begin{table}[t]
\scriptsize
\centering
\caption{Cumulative localisation recall (\%) at angular in degrees and translation thresholds. Cell colour indicates recall level: darker green indicates higher recall.}
\label{tab:recall}
\begin{tabular}{ll cccccc c ccccccc}
\hline
& & \multicolumn{6}{c}{Angular (\%)} & & \multicolumn{7}{c}{Translation (\%)} \\
\cline{3-8} \cline{10-16}
Scene & Method & 5° & 15° & 30° & 45° & 60° & 90° & & 0.5 & 1.0 & 2.5 & 4.0 & 5.0 & 7.5 & 10.0 \\
\hline
\multirow{6}{*}{Arc}
& No DT         & \heatcell{3.8} & \heatcell{34.6} & \heatcell{50.0} & \heatcell{57.7} & \heatcell{57.7} & \heatcell{57.7} && \heatcell{0.0} & \heatcell{0.0} & \heatcell{0.0}  & \heatcell{15.4} & \heatcell{23.1} & \heatcell{73.1} & \heatcell{80.8} \\
& Img. CycleGAN & \heatcell{0.0} & \heatcell{19.2} & \heatcell{26.9} & \heatcell{34.6} & \heatcell{38.5} & \heatcell{46.2} && \heatcell{0.0} & \heatcell{0.0} & \heatcell{0.0}  & \heatcell{11.5} & \heatcell{26.9} & \heatcell{65.4} & \heatcell{80.8} \\
& Img. CUT      & \heatcell{3.8} & \heatcell{15.4} & \heatcell{26.9} & \heatcell{34.6} & \heatcell{34.6} & \heatcell{42.3} && \heatcell{0.0} & \heatcell{0.0} & \heatcell{0.0}  & \heatcell{0.0}  & \heatcell{23.1} & \heatcell{61.5} & \heatcell{92.3} \\
& Img. UNIT     & \heatcell{0.0} & \heatcell{15.4} & \heatcell{15.4} & \heatcell{15.4} & \heatcell{23.1} & \heatcell{26.9} && \heatcell{0.0} & \heatcell{0.0} & \heatcell{0.0}  & \heatcell{3.8}  & \heatcell{15.4} & \heatcell{53.8} & \heatcell{92.3} \\
& Img. MUNIT    & \heatcell{0.0} & \heatcell{11.5} & \heatcell{23.1} & \heatcell{23.1} & \heatcell{23.1} & \heatcell{34.6} && \heatcell{0.0} & \heatcell{0.0} & \heatcell{7.7}  & \heatcell{11.5} & \heatcell{11.5} & \heatcell{34.6} & \heatcell{61.5} \\
& Emb. CycleGAN & \heatcell{3.8} & \heatcell{38.5} & \heatcell{57.7} & \heatcell{57.7} & \heatcell{57.7} & \heatcell{69.2} && \heatcell{0.0} & \heatcell{0.0} & \heatcell{3.8}  & \heatcell{19.2} & \heatcell{34.6} & \heatcell{65.4} & \heatcell{88.5} \\
\hline
\multirow{6}{*}{Brandenburg}
& No DT         & \heatcell{0.0} & \heatcell{22.2} & \heatcell{77.8} & \heatcell{94.4} & \heatcell{94.4} & \heatcell{100.0} && \heatcell{0.0} & \heatcell{0.0} & \heatcell{16.7} & \heatcell{55.6} & \heatcell{77.8} & \heatcell{88.9}  & \heatcell{100.0} \\
& Img. CycleGAN & \heatcell{0.0} & \heatcell{22.2} & \heatcell{77.8} & \heatcell{94.4} & \heatcell{94.4} & \heatcell{100.0} && \heatcell{0.0} & \heatcell{0.0} & \heatcell{16.7} & \heatcell{55.6} & \heatcell{66.7} & \heatcell{100.0} & \heatcell{100.0} \\
& Img. CUT      & \heatcell{0.0} & \heatcell{27.8} & \heatcell{66.7} & \heatcell{88.9} & \heatcell{94.4} & \heatcell{100.0} && \heatcell{0.0} & \heatcell{0.0} & \heatcell{0.0}  & \heatcell{50.0} & \heatcell{77.8} & \heatcell{100.0} & \heatcell{100.0} \\
& Img. UNIT     & \heatcell{0.0} & \heatcell{0.0}  & \heatcell{50.0} & \heatcell{94.4} & \heatcell{94.4} & \heatcell{100.0} && \heatcell{0.0} & \heatcell{0.0} & \heatcell{11.1} & \heatcell{33.3} & \heatcell{44.4} & \heatcell{94.4}  & \heatcell{94.4}  \\
& Img. MUNIT    & \heatcell{0.0} & \heatcell{0.0}  & \heatcell{44.4} & \heatcell{77.8} & \heatcell{88.9} & \heatcell{100.0} && \heatcell{0.0} & \heatcell{0.0} & \heatcell{0.0}  & \heatcell{0.0}  & \heatcell{5.6}  & \heatcell{66.7}  & \heatcell{94.4}  \\
& Emb. CycleGAN & \heatcell{0.0} & \heatcell{22.2} & \heatcell{72.2} & \heatcell{94.4} & \heatcell{94.4} & \heatcell{100.0} && \heatcell{0.0} & \heatcell{0.0} & \heatcell{27.8} & \heatcell{50.0} & \heatcell{72.2} & \heatcell{94.4}  & \heatcell{100.0} \\
\hline
\multirow{6}{*}{Duomo}
& No DT         & \heatcell{0.0} & \heatcell{6.7}  & \heatcell{6.7}  & \heatcell{26.7} & \heatcell{33.3} & \heatcell{66.7} && \heatcell{0.0}  & \heatcell{13.3} & \heatcell{80.0} & \heatcell{100.0} & \heatcell{100.0} & \heatcell{100.0} & \heatcell{100.0} \\
& Img. CycleGAN & \heatcell{0.0} & \heatcell{6.7}  & \heatcell{20.0} & \heatcell{26.7} & \heatcell{40.0} & \heatcell{66.7} && \heatcell{13.3} & \heatcell{26.7} & \heatcell{73.3} & \heatcell{100.0} & \heatcell{100.0} & \heatcell{100.0} & \heatcell{100.0} \\
& Img. CUT      & \heatcell{0.0} & \heatcell{6.7}  & \heatcell{13.3} & \heatcell{20.0} & \heatcell{33.3} & \heatcell{46.7} && \heatcell{0.0}  & \heatcell{26.7} & \heatcell{86.7} & \heatcell{100.0} & \heatcell{100.0} & \heatcell{100.0} & \heatcell{100.0} \\
& Img. UNIT     & \heatcell{0.0} & \heatcell{0.0}  & \heatcell{6.7}  & \heatcell{6.7}  & \heatcell{13.3} & \heatcell{46.7} && \heatcell{6.7}  & \heatcell{6.7}  & \heatcell{60.0} & \heatcell{80.0}  & \heatcell{100.0} & \heatcell{100.0} & \heatcell{100.0} \\
& Img. MUNIT    & \heatcell{0.0} & \heatcell{0.0}  & \heatcell{0.0}  & \heatcell{0.0}  & \heatcell{6.7}  & \heatcell{26.7} && \heatcell{0.0}  & \heatcell{0.0}  & \heatcell{33.3} & \heatcell{73.3}  & \heatcell{93.3}  & \heatcell{100.0} & \heatcell{100.0} \\
& Emb. CycleGAN & \heatcell{0.0} & \heatcell{0.0}  & \heatcell{26.7} & \heatcell{33.3} & \heatcell{53.3} & \heatcell{73.3} && \heatcell{0.0}  & \heatcell{26.7} & \heatcell{93.3} & \heatcell{93.3}  & \heatcell{100.0} & \heatcell{100.0} & \heatcell{100.0} \\
\hline
\end{tabular}
\end{table}

Table~\ref{tab:results_all} reports the results where HistReNeRF achieves the lowest translation error in all three scenes. On Arc de Triomphe, embedding-space adaptation produces a small reduction in MTE from $6.624$m to $6.608$m and reduces MAE from $71.7^\circ$ to $69.6^\circ$. The modest translation gain is consistent with the substantial viewpoint mismatch and rotational symmetry of this scene. On Brandenburg Gate, HistReNeRF reduces MTE by $17\%$, from $4.712$m to $3.912$m, while retaining the baseline MAE of $23.4^\circ$. The greatest improvement occurs on Piazza del Duomo, where MTE decreases from $1.783$,m to $1.589$,m and MAE decreases from $85.5^\circ$ to $58.6^\circ$. In contrast, pixel-space translation does not provide a comparable improvement. Across Arc de Triomphe and Piazza del Duomo, all pixel-space methods increase translation error relative to no adaptation. CUT produces a small translation improvement on Brandenburg Gate, reducing MTE from $4.712$,m to $4.543$,m, but increases MAE from $23.4^\circ$ to $28.2^\circ$. HistReNeRF is therefore the only adaptation approach that improves or preserves both pose measures consistently across the three scenes. The following subsection examines the scene-dependent behaviour of pixel-space adaptation and the factors underlying this difference.

Table~\ref{tab:recall} provides complementary cumulative recall results at angular and translation thresholds. These results show that the benefit of embedding-space adaptation is not limited to mean error. On Arc de Triomphe, angular recall at $90^\circ$ increases from $57.7\%$ to $69.2\%$. On Piazza del Duomo, translation recall at $2.5$,m increases from $80.0\%$ to $93.3\%$, while angular recall at $30^\circ$ increases from $6.7\%$ to $26.7\%. $

The magnitude of improvement depends on contemporary scene coverage. Arc de Triomphe exhibits the smallest gain from embedding-space adaptation and the strongest degradation from pixel-space translation. As shown in Fig.~\ref{fig:qualitative_results}, historic queries include distant, central, and elevated viewpoints that are underrepresented in the street-level Mapillary collection. Appearance adaptation can reduce the historic-to-contemporary feature gap but cannot recover architectural evidence that is absent from the contemporary reconstruction. This mismatch is damaging for pixel-level translation, which can introduce modern street-level textures that are inconsistent with the structure visible in historic views.

\section{Conclusion}
\label{sec:conclusion}

We presented HistReNeRF for relocalising historic photographs within contemporary neural scene reconstructions. The method adapts DINOv2 patch embeddings towards a contemporary feature distribution before ray--query matching, leaving the query image and scene representation unchanged. Experiments on HistScene show that embedding-space adaptation achieves the lowest translation error across all three landmarks while improving or preserving angular accuracy; pixel-space translation is inconsistent and often degrades localisation. The results indicate that adapting the matching representation is more reliable than translating query pixels under cross-temporal appearance change. Future work could evaluate additional site types, including less-photographed and archaeological sites, compare classical localisation baselines, investigate alternatives such as diffusion, and explore richer archival case studies with broader viewpoint and scene coverage.

\bibliographystyle{splncs04}
\bibliography{refs}
\end{document}